\documentclass[conference]{IEEEtran}
\usepackage[utf8]{inputenc}
\usepackage[T1]{fontenc}
\usepackage{amsmath,amssymb,amsthm}
\usepackage{graphicx}
\usepackage{tabularx,array,booktabs}
\usepackage{tikz}
\usepackage{hhline}
\usetikzlibrary{arrows.meta}
\newtheorem{assumption}{Assumption}
\newtheorem{theorem}{Theorem}
\newtheorem{corollary}{Corollary}
\newtheorem{lemma}{Lemma}
\newtheorem{definition}{Definition}
\newif\ifincludeappendix
\includeappendixtrue
\begin{document}
\title{One-shot Distributed Learning via Weighted Averaging with Server-side Information}
\author{
\IEEEauthorblockN{
Pavle Tepav\v{c}evi\'{c}, Du\v{s}an Jakoveti\'{c}, Milo\v{s} Savi\'{c}
}
\IEEEauthorblockA{
Department of Mathematics and Informatics, Faculty of Sciences, University of Novi Sad, Serbia\\
\{pavle.tepavcevic, dusan.jakovetic, svc\}@dmi.uns.ac.rs
}
}
\maketitle

\begin{abstract}

Distributed learning has drawn considerable attention due to its capability of learning from distributed data without aggregating raw observations. However, frequent exchange of model updates can become the main bottleneck of communication. Therefore, one-shot approaches where each worker fits a local model and sends it to the server to aggregate emerge. It is known that averaging can achieve the so-called pooled-data mean squared error (MSE) order of a hypothetical node that has access to data from the server and all the workers, and performs empirical risk minimization (ERM) on it. However, when the number of workers increases too fast compared to the local sample size, nonzero local estimation bias prevents averaging from achieving this order.

In this paper, we study whether independent server data can overcome this limitation. We consider $m$ workers with $n$ observations each and a server with $p$ observations, all i.i.d., giving $N=mn+p$ observations in total. Under certain regularity conditions, a pooled-data ERM achieves MSE of order $\mathcal O(N^{-1})$. We analyze deterministic nonnegative averaging of convex-loss M-estimators, with equal worker weights and the server's own estimator included. Under our assumptions, including nonzero local bias of order $n^{-\alpha}$, $\alpha\geq1$, sample-size weighted averaging achieves the pooled-data MSE order whenever $m=\mathcal O(n^{2\alpha-1})$. Above this worker scale, it does so if and only if $p=\Omega(nm\cdot m/n^{2\alpha-1})$. Hence, server-only estimation, whose MSE is of order $\mathcal{O}(p^{-1})$, can achieve pooled-data order while sample-size weighting fails. Analytical examples and synthetic experiments illustrate when averaging matches the pooled-data MSE order, complemented by CIFAR-10 experiments comparing aggregation strategies with linear and nonlinear classifiers.

\end{abstract}


\section{Introduction}
\label{sec:introduction}

Smart devices and IoT systems produce data from various locations. Collection of all this data at a central server may be expensive or undesirable due to their large number and privacy concerns~\cite{shi2016edge}. This gives rise to learning techniques that operate on data where it resides. Distributed learning involves cooperation between multiple computing nodes in training~\cite{zhang2013}. Federated learning, a special case of distributed learning, is a framework where devices keep their raw training data and collaborate by sharing local updates of models~\cite{mcmahan2017communication}.

Communication is still a big issue, especially in situations where training involves many rounds of model updates transmission over limited bandwidth channels~\cite{mcmahan2017communication}. The one-shot averaging framework that we consider involves local fitting of models and transmission of their parameters to the server only once~\cite{zhang2013,rosenblatt2016}. It eliminates the need for many communication rounds, but aggregation of local models can suffer from loss of statistical efficiency compared to joint fitting of a model on all data.

We consider $m$ workers with $n$ observations each and a server with an independent dataset of size $p$, that is, there are $N=mn+p$ i.i.d. observations in total. In addition to aggregation of models fitted by the workers, the server fits its own estimator based on its $p$ observations. The benchmark is a hypothetical pooled-data ERM that performs fitting on all $N$ observations. 

\noindent\textbf{Key research question.}
\emph{Is it possible for the independent server data to help one-shot weighted averaging achieve the MSE order of the pooled-data, and what size should the server data be?}

\subsection{Related work}
Zhang et al.~\cite{zhang2013} analyze averaging independently fitted worker M-estimators, without an independent server estimator. Their bound has order $\mathcal{O}((mn)^{-1}+n^{-2})$, guaranteeing the pooled-data MSE order for $m=\mathcal O(n)$ when dimension and regularity constants are fixed. They also consider a bootstrap subsampling-based bias correction which improves on simple averaging achieving MSE of order $\mathcal{O}((mn)^{-1}+n^{-3})$. This approach is more robust to parallelization, achieving the pooled-data MSE order for $m=\mathcal O(n^2)$.

Rosenblatt and Nadler~\cite{rosenblatt2016} establish first-order equivalence between worker averaging and pooled estimation as local sample size grows with both dimension and the number of workers fixed. This does not cover arbitrary growth of $m$ with $n$.

Armacki et al.~\cite{armacki2024oneshot} study one-shot learning with heterogeneous data by clustering local models and averaging within clusters. We instead study i.i.d.\ data with an independent server dataset.

Dobriban and Sheng~\cite{dobriban2020} consider weighted one-shot ridge regression in a high-dimensional random effects model. The weights which minimize their error may sum to a value greater than one due to shrinkage. Our weights are nonnegative and sum to one, and we also examine the sample size of the independent server data.

Trusted server information has already been explored in robust learning. FLTrust~\cite{cao2021fltrust} evaluates and normalizes client updates based on the server dataset, while Zeno~\cite{xie2019zeno} uses the server-side loss estimate to rank the gradients and FedGreed~\cite{kritharakis2025fedgreed} uses the server-side loss to select the client models. In each of these approaches, the issue of faulty or adversarial clients is handled using iterative training. We instead use server data to fit an estimator included in a one-shot average, under i.i.d. sampling and without adversarial clients.

\subsection{Contributions}

The main contributions are:
\begin{itemize}
    \item We derive the MSE order of sample-size weighting, which assigns weight $n/N$ to each worker estimator and $p/N$ to the server estimator. We establish necessary and sufficient conditions for attaining the pooled-data MSE order, characterizing the required server sample size.

    \item We identify regimes where server-only estimation
    attains the pooled-data MSE order while sample-size weighting
    does not, showing that incorporating worker models can
    worsen the MSE order.

    \item We provide analytical examples distinguishing zero bias
    from bias of orders $n^{-1}$ and $n^{-2}$, and compare
    aggregation strategies across sample-size regimes through
    synthetic and CIFAR-10 experiments.
\end{itemize}

\par\smallskip
\noindent\textbf{Paper organization.}
Section~\ref{sec:setup} describes the model and the assumptions. The results are presented in Section~\ref{sec:results}. Sections~\ref{sec:concrete_examples} and~\ref{sec:numerics} give analytical and numerical illustrations.
\ifincludeappendix
The appendix contains the proofs that are not included in the main text.
\fi
The rest of this section introduces the notation used in this paper.

\par\smallskip
\noindent\textbf{Notation.}
Notations $\|\cdot\|_2$ and $\|\cdot\|_F$ stand for Euclidean/spectral and Frobenius norms; $\langle\cdot,\cdot\rangle$ is the Euclidean inner product. Notation $A\succeq B$ is used for positive semidefiniteness of matrix $A-B$; $A\succ0$ is positive definiteness of $A$; $I_d$ is the $d\times d$ identity matrix. Notations $\nabla$, $\nabla^2$, $\operatorname{Cov}$, and $\operatorname{int}$ stand for gradient, Hessian, covariance, and interior, respectively. For two nonnegative sequences $\{a_n\}_{n\in\mathbb{N}}$ and $\{b_n\}_{n\in\mathbb{N}}$, notation $a_n=\mathcal{O}(b_n)$ ($a_n=\Omega(b_n)$) means that there exists a constant $C_1>0$ and $n_1\in\mathbb{N}$ such that $a_n\leq C_1b_n$ ($a_n\geq C_1b_n$) holds for all $n\geq n_1$. The relations $a_n=o(b_n)$ and $a_n=\omega(b_n)$ are defined by the requirement that the above inequalities hold for any $C_1>0$, with $n_1$ depending on $C_1$. We define $\Theta(b_n):=\mathcal O(b_n)\cap\Omega(b_n)$.

\section{Problem Formulation and Assumptions}
\label{sec:setup}
Let $X\sim P$ and let $f(\theta;x)$ be convex in $\theta\in\Theta\subseteq\mathbb R^d$, with $\mathbb E|f(\theta;X)|<\infty$. We assume that the population risk has a unique minimizer:
\begin{equation}
\theta^*=\arg\min_{\theta\in\Theta}F_0(\theta),
\qquad F_0(\theta)=\mathbb E[f(\theta;X)].
\label{eq:population}
\end{equation}

An M-estimator $\hat\theta_k$ is obtained by empirical risk minimization (ERM), which minimizes the average loss over $k$ independent observations. Each worker computes:
\begin{equation}
\theta_i\in\arg\min_{\theta\in\Theta}
\frac1n\sum_{j=1}^{n}f(\theta;X_{ij}),
\qquad i=1,\ldots,m.
\label{eq:local_estimators}
\end{equation}
If $p>0$, the server computes:
\begin{equation}
\theta_0\in\arg\min_{\theta\in\Theta}
\frac1p\sum_{j=1}^{p}f(\theta;X_{0j}).
\label{eq:server_estimator}
\end{equation}
When $p=0$, the server estimator is omitted. All the data points are independent samples from distribution $P$. Since $f(\cdot;x)$ and $\Theta$ are convex (see Assumption \ref{ass:1} ahead), each node has to solve a convex optimization problem based on its data points alone, and hence any local minimum is a global minimum.

Each worker sends its parameter vector once. The server computes its own estimator and forms
\begin{equation}
\bar\theta=c_1\sum_{i=1}^m\theta_i+c_0\theta_0,
\quad c_0,c_1\geq0,\quad mc_1+c_0=1.
\label{eq:aggregation}
\end{equation}
Weights are deterministic and may depend on $m,n,p$. We compare two choices:

\begin{table}[htbp]
\centering
\caption{Averaging strategies, where $N=mn+p$.}
\label{tab:strategies}
\renewcommand{\arraystretch}{1.2}
\begin{tabular}{|l||c|c|}
\hline
Strategy & $c_0$ & $c_1$ \\
\hhline{|=||=|=|}
Server only ($p>0$) & $1$ & $0$ \\
\hline
Sample-size weighting & $p/N$ & $n/N$ \\
\hline
\end{tabular}
\end{table}

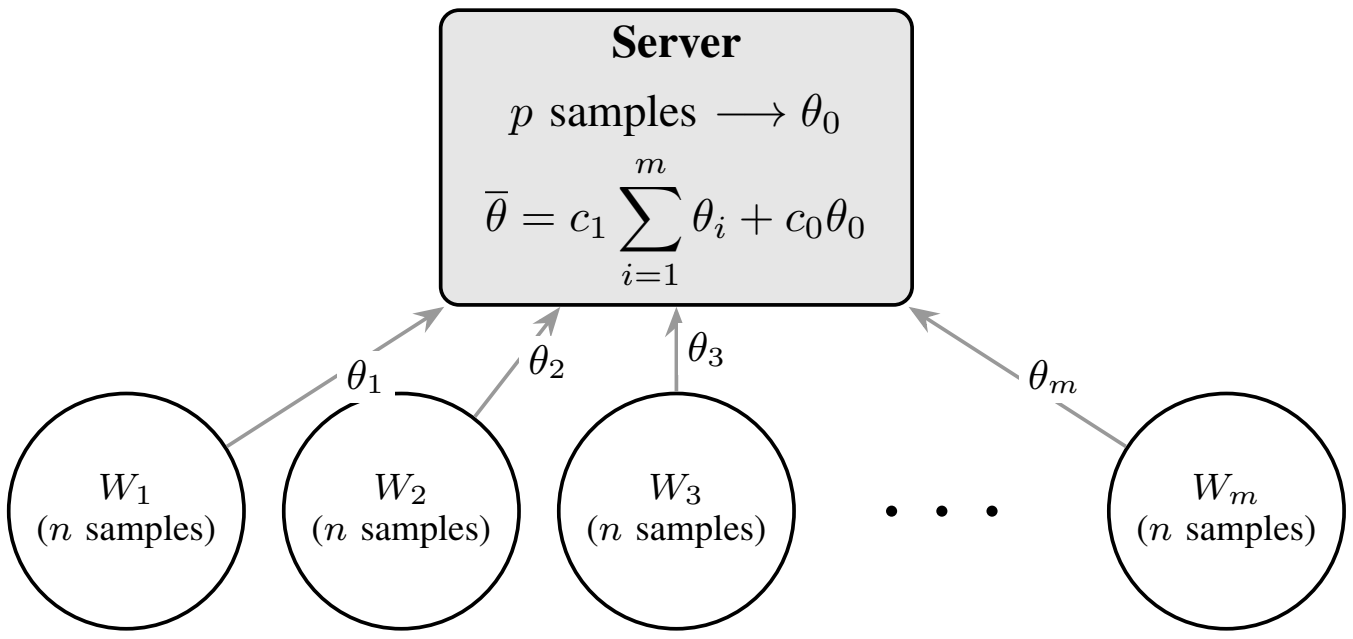
\begin{figure}[htbp]
 \centering
 \resizebox{\linewidth}{!}{%
 \begin{tikzpicture}[
 server/.style={
 rectangle, rounded corners=4pt,
 draw=black, fill=gray!20, thick,
 minimum width=3.6cm, minimum height=1.4cm,
 align=center, font=\normalsize
 },
 client/.style={
 circle, draw=black, fill=white, thick,
 minimum size=1.0cm,
 align=center, font=\footnotesize
 },
 comm/.style={->, >=Stealth, thick, gray!80},
 estimate/.style={
 midway, right, text=black,
 fill=white, inner sep=2pt, font=\small
 }
 ]
 \node[server] (S) at (0, 0) {
 \textbf{Server}\\[3pt]
 $p$ samples $\longrightarrow \theta_0$\\[4pt]
 $\displaystyle
 \overline{\theta}
 =c_1\sum_{i=1}^{m}\theta_i+c_0\theta_0$
 };

 \node[client] (W1) at (-4.2, -2.7) {$W_1$\\($n$ samples)};
 \node[client] (W2) at (-2.1, -2.7) {$W_2$\\($n$ samples)};
 \node[client] (W3) at (0, -2.7) {$W_3$\\($n$ samples)};
 \node[font=\Huge] (dots) at (2.1, -2.7) {$\dots$};
 \node[client] (Wm) at (4.2, -2.7) {$W_m$\\($n$ samples)};

 \draw[comm] (W1) -- (S) node[estimate] {$\theta_1$};
 \draw[comm] (W2) -- (S) node[estimate] {$\theta_2$};
 \draw[comm] (W3) -- (S) node[estimate] {$\theta_3$};
 \draw[comm] (Wm) -- (S) node[estimate] {$\theta_m$};
 \end{tikzpicture}%
 }
 \caption{The workers send their local estimates to the server, which combines them with its own estimate to obtain $\overline{\theta}$.}
 \label{fig:distributed_setup}
\end{figure}

The server forms a weighted average of the worker estimators
and, when $p>0$, its own estimator
(Figure~\ref{fig:distributed_setup}).

The hypothetical pooled estimator has access to every raw observation:
\begin{equation}
\hat\theta_{\mathrm{pool}}\in\arg\min_{\theta\in\Theta} \frac1N\left( \sum_{i=1}^{m}\sum_{j=1}^{n}f(\theta;X_{ij}) +\sum_{j=1}^{p}f(\theta;X_{0j}) \right),
\label{eq:pooled}
\end{equation}
The MSE of $\hat\theta_{\mathrm{pool}}$ with respect to $\theta^*$ under Assumptions~\ref{ass:1}--\ref{ass:3} is of order $\mathcal O(N^{-1})$~\cite[Lemma~8]{zhang2013}. This pooled-data rate defines our benchmark.

\begin{definition}[Order-optimality]
\label{def:order_optimality}
A strategy with estimator $\bar\theta$ is \emph{order-optimal}
if:
\[
\mathbb E\|\bar\theta-\theta^*\|_2^2 =\mathcal O\!\left(\frac1{N}\right).
\]
\end{definition}

Therefore, order optimality is equivalent to matching the MSE rate of the pooled data. As an example, both $2/N$ and $10/N$ are order-optimal, regardless of whether their constant terms match. This definition does not imply that the weights must be such that the MSE is minimized for any given sample size or that they are optimal among all estimators.

We make the following assumptions:

\begin{assumption}[Parameters]
\label{ass:1}
The set $\Theta\subset\mathbb R^d$ is compact and convex, with $\theta^*\in\operatorname{int}\Theta$ and radius $R=\max_{\theta\in\Theta}\|\theta-\theta^*\|_2$.
\end{assumption}

\begin{assumption}[Local strong convexity]
\label{ass:2}
The population risk is twice differentiable and $\nabla^2F_0(\theta^*)\succeq\lambda I_d$ for some $\lambda>0$.
\end{assumption}

\begin{assumption}[Smoothness]
\label{ass:3}
Let $U=\{\theta:\|\theta-\theta^*\|_2\leq\rho\}\subseteq\Theta$ for some $\rho>0$. First and second derivatives of $f$ exist on $U$, and finite constants $G,H$ satisfy
\[
\begin{split}
\mathbb E\|\nabla f(\theta;X)\|_2^8&\leq G^8,\\
\mathbb E\|\nabla^2f(\theta;X)-\nabla^2F_0(\theta)\|_2^8&\leq H^8
\end{split}
\]
for all $\theta\in U$. For every $x$ and $\theta,\theta'\in U$,
\[
\|\nabla^2f(\theta';x)-\nabla^2f(\theta;x)\|_2 \leq L(x)\|\theta'-\theta\|_2.
\]
For some finite $L$, $\mathbb E[L(X)^8]\leq L^8$ and $\mathbb E[(L(X)-\mathbb E L(X))^8]\leq L^8$.
\end{assumption}

\begin{assumption}[Gradient covariance]
\label{ass:4}
The covariance matrix $K=\operatorname{Cov}(\nabla f(\theta^*;X))$ is finite and positive definite.
\end{assumption}

\begin{assumption}[Nonzero asymptotic bias]
\label{ass:5}
For the $k$-sample empirical risk minimizer $\hat\theta_k$ defined above, there exist $\alpha\geq1$ and a constant vector $b\neq0$ such that:
\[
\lim_{k\to\infty}k^\alpha\mathbb E[\hat\theta_k-\theta^*]=b.
\]
\end{assumption}

Assumptions~\ref{ass:1}--\ref{ass:3} are the regularity conditions used in Zhang et al.~\cite{zhang2013}; Assumption~\ref{ass:4} does not allow degenerate gradient noise. 

Assumption~\ref{ass:5} restricts our consideration to the cases where the local estimators have a nonzero leading bias term of order $k^{-\alpha}$, with $\alpha\geq1$. So, it excludes unbiased estimators. Relaxing the requirement $\alpha\geq 1$ would not create new admissible models, because the bias bound in the proof of Theorem~1 of Zhang et al.~\cite[Appendix~B.0.2]{zhang2013}, under Assumptions~\ref{ass:1}--\ref{ass:3}, gives
\[
\left\|\mathbb{E}[\hat{\theta}_k-\theta^*]\right\|_2=O(k^{-1}).
\]
Hence, for any fixed $\alpha<1$, it follows that
\[
k^\alpha
\left\|\mathbb{E}[\hat{\theta}_k-\theta^*]\right\|_2
=O(k^{\alpha-1})
\longrightarrow 0.
\]
This contradicts the requirement $b\neq 0$. That is why it is enough to keep $\alpha\geq 1$ in Assumption~\ref{ass:5}. Section~\ref{sec:concrete_examples} presents examples for this assumption with $\alpha=1$ and $\alpha=2$, as well as an example when this assumption does not hold showing why the nonzero-bias condition is needed for the necessity results for sample-size weighting.

In order to study the impact of data partitioning alone, we fix the distribution, loss function, dimensionality, and the constants in the assumptions. All asymptotic statements take $n\to\infty$, with integer sequences $m=m(n)\geq2$ and $p=p(n)\geq0$. We consider stable scaling regimes in which
\[
p,\qquad \frac{m}{n^{2\alpha-1}},\qquad \frac{p}{mn},\qquad \frac{pn^{2\alpha-2}}{m^2}
\]
each have a limit in $[0,+\infty]$. The presence of such a condition helps to classify asymptotic regimes by avoiding oscillatory scaling; all our results are stated under this assumption.

The convergence of these sequences guarantees that the system is in some well-defined asymptotic regime. For example, convergence of $m/n^{2\alpha-1}$ separates the cases $m=\mathcal O(n^{2\alpha-1})$ and $m=\omega(n^{2\alpha-1})$. Without this restriction, a sequence may alternate between regimes and require separate conclusions along different subsequences. Growth does not need to be monotone; it is allowed to have fluctuations, as long as the four sequences above keep converging to their respective limits.

\section{Results}
\label{sec:results}
We first characterize sample-size weighting, then compare it with server-only estimation and explain the bias mechanism.

\subsection{Sample-size weighting}
The following theorem applies to the local estimators in~\eqref{eq:local_estimators} and the average in~\eqref{eq:aggregation}. Recall that order-optimality (Definition~\ref{def:order_optimality}) means achieving MSE $\mathcal O(N^{-1})$.

\begin{theorem}[Sample-weighted strategy]
\label{thm:sample-weighted-strategy}
Under Assumptions~\ref{ass:1}--\ref{ass:5}, let $c_0=p/N$ and $c_1=n/N$. Then
\begin{equation}
\operatorname{MSE}(\bar\theta) = \Theta\!\left(\frac1N+\frac{m^2}{N^2n^{2\alpha-2}}\right).
\label{eq:sample_weight_mse}
\end{equation}
Sample-size weighting is order-optimal if and only if
$m^2/n^{2\alpha-2}=\mathcal O(N)$.
\end{theorem}

\begin{corollary}[Server sample-size requirement]
\label{cor:sample-weighted-regimes}
Under Assumptions~\ref{ass:1}--\ref{ass:5}:
\begin{enumerate}
\item If $m=\mathcal O(n^{2\alpha-1})$, sample-size weighting is order-optimal for every $p\geq0$.
\item If $m=\omega(n^{2\alpha-1})$, it is order-optimal if and only if $p=\Omega(nm\cdot m/n^{2\alpha-1})$.
\end{enumerate}
\end{corollary}

\begin{corollary}[Server-only versus sample-size weighting]
\label{cor:gap}
Under Assumptions~\ref{ass:1}--\ref{ass:5}, if
\[
m=\omega(n^{2\alpha-1}),\quad
p=\Omega(nm),\quad
p=o(nm\cdot m/n^{2\alpha-1}),
\]
server-only estimation is order-optimal, whereas sample-size weighting is not.
\end{corollary}

\subsection{Implications for weighted averaging}
Rewriting~\eqref{eq:sample_weight_mse} gives:
\[
\operatorname{MSE}(\bar\theta)=\Theta\!\left(\frac1N+\left(\frac{mn}{N}\right)^2\frac1{n^{2\alpha}}\right).
\]
The factor $mn/N$ is the total worker weight. More server data reduce the workers' bias contribution, but this reduction must keep pace with the improving $N^{-1}$ benchmark. Table~\ref{tab:averaging_regimes} summarizes the two strategies.

\begin{table}[!htbp]
\centering
\caption{Necessary and sufficient conditions for order-optimality, under
Assumptions~\ref{ass:1}--\ref{ass:5}.}
\label{tab:averaging_regimes}
\small
\setlength{\tabcolsep}{3pt}
\renewcommand{\arraystretch}{1.15}
\begin{tabularx}{\columnwidth}{|>{\raggedright\arraybackslash}X||c|c|}
\hline
Strategy
& $m=\mathcal O(n^{2\alpha-1})$
& $m=\omega(n^{2\alpha-1})$\\
\hhline{|=||=|=|}
Server only
& $p=\Omega(mn)$
& $p=\Omega(mn)$\\
\hline
Sample-size weights \vspace{0.5mm}
& Any $p\geq0$
& $p=\Omega\!\left(nm\cdot\frac{m}{n^{2\alpha-1}}\right)$\\
\hline
\end{tabularx}
\end{table}

When $m=\omega(n^{2\alpha-1})$, the required server sample-size scale for sample-size weighting is larger than that for server-only estimation by the diverging factor $m/n^{2\alpha-1}$. For example, if $p=\Theta(mn)$, the server stores a constant fraction of all samples, which makes the server-only estimation order-optimal. In contrast, sample-size weighting provides a non-vanishing total weight to the worker estimators. This leads to the contribution of the common local bias into the squared error of the higher order than $N^{-1}$, thus making it impossible for order-optimality.

The above statements apply to the averaging procedures considered in this paper; other one-shot aggregation procedures could be more successful. Having several rounds of communication, it is possible, for example, to implement FedAvg~\cite{mcmahan2017communication} under certain convergence assumptions in order to approximate the hypothetical ERM procedure using the information provided by both workers and the server. Therefore, the above results describe an inefficiency of sample-size weighting in a one-shot framework without claiming the ineffectiveness of federated learning.

\ifincludeappendix
Proofs and further details are given in Appendices~\ref{app:local} and~\ref{app:weighting}.
\else
Detailed proofs are omitted due to space constraints and will be included in an extended version of this work.
\fi

\section{Insights through Analytical Examples}
\label{sec:concrete_examples}
These scalar examples permit explicit calculations of local estimators and their bias. Although simple, they clarify the role of Assumption~\ref{ass:5} in the results of Section~\ref{sec:results} and show how the bias exponent determines the worker and server sample-size requirements for order-optimality. Table~\ref{tab:loss_assumptions} records the assumptions they satisfy.

\begin{table}[!htbp]
\centering
\caption{Assumptions satisfied by the analytical examples.}
\label{tab:loss_assumptions}
\small
\setlength{\tabcolsep}{3pt}
\renewcommand{\arraystretch}{1.15}
\begin{tabularx}{\columnwidth}{|>{\raggedright\arraybackslash}X||c|c|c|}
\hline
Setting
& \shortstack{Assumptions\\\ref{ass:1}--\ref{ass:4}}
& \shortstack{Assumption\\\ref{ass:5}}
& $\alpha$\\
\hhline{|=||=|=|=|}
Bernoulli sample mean with squared loss & Yes & No & ---\\
\hline
Misspecified least squares & Yes & Yes & $1$\\
\hline
Squared loss with a quartic penalty & Yes & Yes & $2$\\
\hline
\end{tabularx}
\end{table}

\subsection{Bernoulli sample mean with squared loss}
\label{sec:unbiased_mean}
Let $X\sim\operatorname{Bernoulli}(q)$, $0<q<1$, and $f(\theta;X)=(\theta-X)^2/2$ on $\Theta=[-1,2]$. The population minimizer is $\theta^*=q$ and the empirical-risk minimizer is $\hat\theta_n=\bar X_n$. It is unbiased, and sample-size weighting equals pooled ERM exactly, with MSE $q(1-q)/N$ for every number of workers. Thus Assumptions~\ref{ass:1}--\ref{ass:4} do not imply Assumption~\ref{ass:5}. Corollary~\ref{cor:sample-weighted-regimes} does not apply to this example: without local bias, sample-size weighting can remain order-optimal for arbitrarily many workers, regardless of the server sample size.

\subsection{Misspecified least squares}
\label{sec:misspecified_least_squares}
Let $Z$ equal $1$ or $2$ with equal probability, $Y=Z^3$, and $f(\theta;(Z,Y))=(\theta Z-Y)^2/2$ on $\Theta=[0,5]$. No single $\theta$ represents the conditional mean $Z^3$ at both values of $Z$. The minimizers and bias are
\[
\theta^*=\frac{17}{5},\qquad \hat\theta_n=\frac{\sum_{j=1}^{n}Z_j^4}{\sum_{j=1}^{n}Z_j^2},
\]
\[
\mathbb E[\hat\theta_n-\theta^*] =-\frac{72}{125n}+\mathcal O(n^{-3/2}).
\]
Here $\alpha=1$, so Corollary~\ref{cor:sample-weighted-regimes} gives a critical worker scale of $n$ for sample-size weighting and requires $p=\Omega(m^2)$ when $m=\omega(n)$. In particular, $m=\omega(n)$ and $p=\Theta(mn)$ yield the gap identified in Section~\ref{sec:results}: server-only estimation is order-optimal, whereas sample-size weighting is not. This example shows that even a squared-loss problem can exhibit the server-only advantage because its local estimators are biased relative to the population minimizer.

\subsection{Squared loss with a quartic penalty}
\label{sec:quartic_example}
Let $X\in[-1,1]$ have zero mean and nonzero third moment, and set
\[
f(\theta;X)=\tfrac12(\theta-X)^2+\tfrac14\theta^4, \qquad \Theta=[-1,1].
\]
Then $\theta^*=0$, $\hat\theta_n+\hat\theta_n^3=\bar X_n$, and
\[
\mathbb E[\hat\theta_n-\theta^*]=-\frac{\mathbb E[X^3]}{n^2}+\mathcal O(n^{-5/2}).
\]
A concrete choice is $\mathbb P(X=-1/3)=2/3$ and
$\mathbb P(X=2/3)=1/3$, giving $b=-2/27$.

Here $\alpha=2$, so Corollary~\ref{cor:sample-weighted-regimes} guarantees order-optimality of sample-size weighting for any $p\geq0$ when $m=\mathcal O(n^3)$. When $m=\omega(n^3)$, it is order-optimal if and only if $p=\Omega(m^2/n^2)$. Compared with the previous example, faster bias decay raises the critical worker scale from $n$ to $n^3$, permitting more workers without an additional server sample-size requirement.

\ifincludeappendix
The calculations and assumption checks are in Appendix~\ref{app:examples}.
\fi

\section{Numerical Experiments}
\label{sec:numerics}
We compare sample-size weighting, server-only estimation, and pooled fitting across sample-size regimes, focusing on aggregation rather than absolute classification accuracy.

\subsection{Synthetic examples: bias and server sample size}
\begin{figure*}[!t]
\centering
\includegraphics[width=\textwidth]{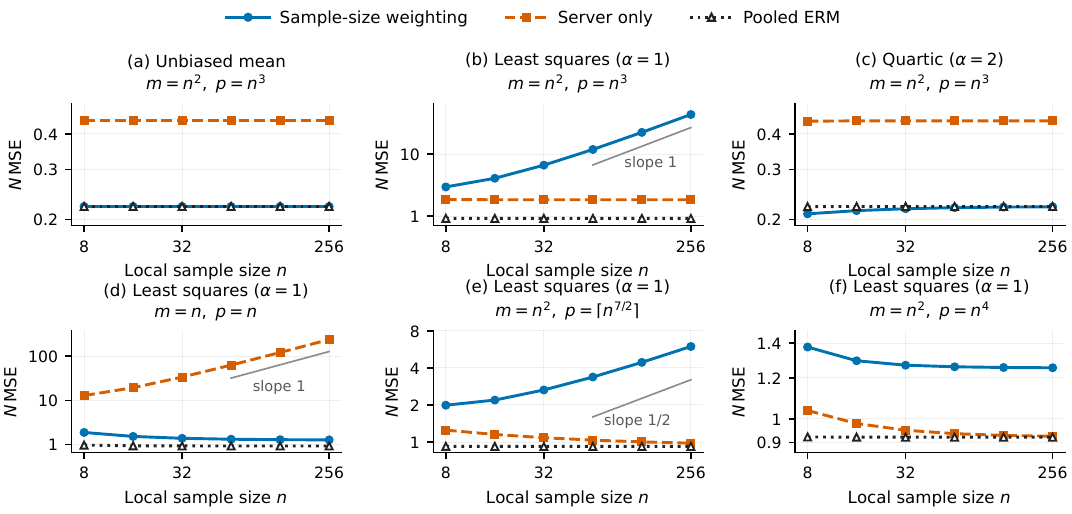}
\caption{Scaled MSE for the analytical examples. Gray lines indicate predicted growth rates. In (a), sample-size weighting and pooled ERM coincide. Vertical scales differ.}
\label{fig:num_synthetic}
\end{figure*}

We use the models of Section~\ref{sec:concrete_examples}, with $q=1/3$ for the Bernoulli mean and the stated quartic distribution. For $n\in\{8,16,32,64,128,256\}$, local moments are computed from binomial statistics, omitting tail probability at most $10^{-18}$. Independence gives the aggregation MSE. These are numerical expectations, not Monte Carlo estimates. Figure~\ref{fig:num_synthetic} reports $N\operatorname{MSE}$, which remains bounded for order-optimal strategies as $n\to\infty$.

Panels~(a)--(c) fix $m=n^2$, $p=n^3$. Sample-size weighting equals pooled ERM for the unbiased mean, loses order-optimality for least squares ($\alpha=1$), and retains it for the quartic model ($\alpha=2$). Server-only estimation is order-optimal throughout; the least-squares gap illustrates Corollary~\ref{cor:gap}.

For least squares with $m=n^2$, Theorem~\ref{thm:sample-weighted-strategy} predicts scaled MSE of orders $n$, $n^{1/2}$, and $1$ for $p=n^3$, $p=\lceil n^{7/2}\rceil$, and $p=n^4$ (panels~(b), (e), (f)). Panel~(d), with $m=p=n$, instead shows order-optimal sample-size weighting and growing server-only scaled MSE.

\subsection{CIFAR-10: linear and nonlinear classifiers}
CIFAR-10~\cite{krizhevsky2009} images use unit-norm, 512-dimensional features from a frozen ImageNet-pretrained ResNet-18~\cite{he2016resnet}. Worker and server data are sampled independently with replacement from the 50,000 training images, with identical draws across methods within each repetition. The 10,000 test images are used only for evaluation.

Ridge uses mean half squared loss on one-hot labels with penalty $10^{-2}\|W\|_F^2/2$ and is solved directly. Linear and nonlinear cross-entropy heads use penalty $10^{-4}\|\theta\|_2^2/2$ and 400 full-batch Adam updates with learning rate $0.01$. The nonlinear head is $512\!\to\!128\!\to\!10$ with a ReLU hidden layer. For each cross-entropy head and repetition, local and pooled fits share the initialization and training budget. Aggregation averages fitted parameters.

Ridge parameter MSE uses the ridge minimizer on the full training pool as reference, distinct from the pooled fit on sampled data. Results average 20 repetitions for ridge and 10 per cross-entropy head; bands are pointwise 95\% Student-$t$ confidence intervals for the means.

\begin{figure*}[!t]
\centering
\includegraphics[width=\textwidth]{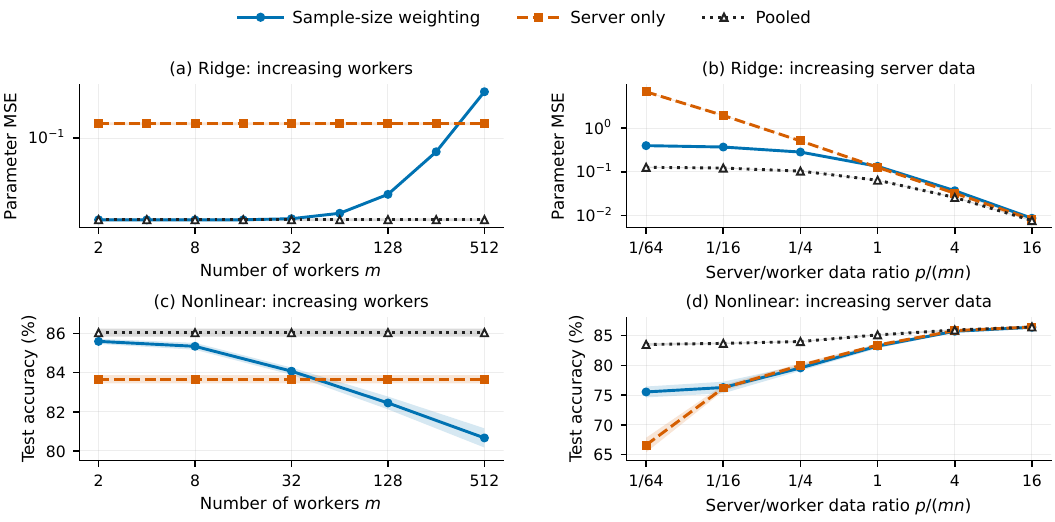}
\caption{CIFAR-10: ridge parameter MSE (top) and nonlinear-head test accuracy (bottom). Left: $mn=32768$, $p=8192$, varying $m$. Right: $mn=8192$, $m=128$, varying $p$. Bands show 95\% confidence intervals.}
\label{fig:num_cifar}
\end{figure*}

\emph{Increasing workers.}
With $mn=32768$, $p=8192$, increasing $m$ from 2 to 512 causes ridge sample-size-weighted MSE to surpass server-only MSE (Figure~\ref{fig:num_cifar}(a)). Nonlinear weighted accuracy falls from $85.59\%$ to $80.66\%$ (panel~(c)); at $m=512$, server-only accuracy is $2.99$ percentage points higher (95\% paired confidence interval: $[2.56,3.42]$). Linear cross-entropy likewise favors server-only estimation at $m=512$: $83.14\%$ versus $80.06\%$ weighted accuracy.

\emph{Increasing server data.}
Panels~(b) and~(d) fix $mn=8192$, $m=128$, and vary $p/(mn)$ from $1/64$ to $16$. Ridge sample-size weighting initially outperforms server-only estimation, but the ordering reverses as $p$ grows. Nonlinear weighted accuracy rises from $75.53\%$ to $86.38\%$, approaching pooled accuracy $86.44\%$.

Doubling training to 800 updates in three repetitions preserves the mean nonlinear server-only advantage at $m=512$, without certifying convergence. We establish neither Assumption~\ref{ass:5} nor a bias exponent for CIFAR-10. The nonlinear head lies outside the convex theory; optimization and parameter alignment may affect its results.

\section{Conclusion}
\label{sec:conclusion}

Under the stated assumptions, local estimation bias determines the important worker scale $n^{2\alpha-1}$. When $m=\mathcal O(n^{2\alpha-1})$, sample-size weighting achieves the pooled-data MSE order for every server sample size. When $m=\omega(n^{2\alpha-1})$, it does so if and only if $p=\Omega(nm\cdot m/n^{2\alpha-1})$. That means that including worker estimators through sample-size weighting can prevent order-optimality even when server-only estimation achieves it.

Synthetic experiments illustrate the predicted regimes and the role of the bias exponent. CIFAR-10 experiments show related behavior: sample-size weighting can underperform server-only estimation with both linear and nonlinear classifiers. The nonlinear results illustrate the relevance of this comparison beyond convex models, but do not establish an extension of the theoretical guarantees.

These conclusions concern the specified one-shot averaging rules under i.i.d.\ sampling and the stated regularity and bias assumptions. They do not exclude better aggregation rules or methods using multiple communication rounds. Heterogeneous data and computational costs remain outside the analysis.

\section*{Acknowledgment}
This paper gives a subset of content from Pavle Tepav\v{c}evi\'{c}'s master thesis, \emph{One-shot Distributed Learning with Server-side Information}, completed in 2026 at the Faculty of Sciences, University of Novi Sad, under the supervision of professor Du\v{s}an Jakoveti\'{c}. The authors acknowledge funding from the Horizon Europe programme through the TURING project under grant agreement No. 101215032.

\ifincludeappendix

\clearpage

\onecolumn

\appendices

\numberwithin{theorem}{section}

\numberwithin{corollary}{section}

\numberwithin{lemma}{section}

\numberwithin{equation}{section}

\section{MSE and Bias Bounds for Local Estimators}
\label{app:local}

All statements below use the notation and asymptotic setting of Section~\ref{sec:setup}. For $n$ i.i.d.\ observations, we write:
\[
F_n(\theta)=\frac1n\sum_{i=1}^n f(\theta;X_i), \qquad \hat\theta_n\in\arg\min_{\theta\in\Theta}F_n(\theta).
\]

\begin{theorem}[Squared Bias Bounds for M-estimators]
\label{thm:boundLocalSquaredBias}
Under Assumption~\ref{ass:5},
\[
\|\mathbb{E}[\hat{\theta}_n-\theta^*]\|_2^2=\Theta(n^{-2\alpha}).
\]
\end{theorem}
\begin{proof}
From Assumption~\ref{ass:5} we have:
\[
(\exists b\neq 0) \quad n^\alpha \, \mathbb{E}\bigl[\hat{\theta}_n - \theta^*\bigr] \to b \qquad \text{as } n \to \infty.
\]
Hence:
\begin{align*}
(\forall\varepsilon>0)(\exists N\in\mathbb{N})(\forall n\in\mathbb{Z}_{\geq N})\quad\|n^\alpha \, &\mathbb{E}\bigl[\hat{\theta}_n - \theta^*\bigr] - b\|_2 \leq \varepsilon. \\
\|b\|_2 - \varepsilon &\leq \|n^\alpha \, \mathbb{E}\bigl[\hat{\theta}_n - \theta^*\bigr]\|_2 \leq \|b\|_2 + \varepsilon. \\
\bigl(\|b\|_2 - \varepsilon\bigr) n^{-\alpha} &\leq \bigl\|\mathbb{E}\bigl[\hat{\theta}_n - \theta^*\bigr]\bigr\|_2 \leq \bigl(\|b\|_2 + \varepsilon\bigr) n^{-\alpha}.
\end{align*}
Now let \(\varepsilon = \|b\|_2 / 2\). From \(b \neq 0\), we have \(\|b\|_2 > 0\).
\[
\bigl(\|b\|_2 - \varepsilon\bigr)^2 n^{-2\alpha} \leq \bigl\|\mathbb{E}\bigl[\hat{\theta}_n - \theta^*\bigr]\bigr\|_2^2 \leq \bigl(\|b\|_2 + \varepsilon\bigr)^2 n^{-2\alpha}.
\]
For all \(n\) big enough,
\[
\frac{\|b\|_2^2}{4} n^{-2\alpha}
\leq
\bigl\|\mathbb{E}\bigl[\hat{\theta}_n - \theta^*\bigr]\bigr\|_2^2
\leq
\frac{9\|b\|_2^2}{4} n^{-2\alpha}.
\]
Then by definition of $\Theta$ we have:
\[
\bigl\|\mathbb{E}\bigl[\hat{\theta}_n - \theta^*\bigr]\bigr\|_2^2
=
\Theta\bigl(1 / n^{2\alpha}\bigr).
\]
\end{proof}

Assumptions \ref{ass:1}-\ref{ass:3} are the same as Assumptions 1-3 in \cite{zhang2013}. That means that plugging in $k = 2$ in Lemma 8 from that paper yields the following theorem:

\begin{theorem}[Upper bound of MSE for M-estimators]
\label{thm:upperboundLocalMSE}
Under Assumptions~\ref{ass:1}--\ref{ass:3},
\[
\mathbb{E}[\|\hat{\theta}_n-\theta^*\|_2^2]
=\mathcal O(n^{-1}).
\]
\end{theorem}

To prove Theorem \ref{thm:lowerboundLocalMSE} some well known theorems will be needed. Here, we just state them. Their proofs can be found in the corresponding literature.

\begin{lemma}[Dominated Convergence Theorem for Differentiation] \label{lem:dct_diff}
    Let $U \subset \mathbb{R}^d$ be an open set, and let $f: \mathcal{X} \times U \to \mathbb{R}$ be a function such that $x \mapsto f(x, \theta)$ is integrable for each $\theta \in U$. Suppose that the gradient $\nabla_\theta f(x, \theta)$ exists for almost all $x \in \mathcal{X}$ and for all $\theta \in U$, and there exists an integrable function $g: \mathcal{X} \to \mathbb{R}_{\geq 0}$ (i.e., $\mathbb{E}[g(X)] < \infty$) such that $\|\nabla_\theta f(x, \theta)\|_2 \leq g(x)$ for all $\theta \in U$. Then, the expectation $\mathbb{E}[f(X, \theta)]$ is differentiable, and the derivative and expectation can be exchanged \cite[multidimensional generalization of Theorem 2.27(b)]{folland1999real}:
    \[
        \nabla_\theta \mathbb{E}[f(X, \theta)] = \mathbb{E}[\nabla_\theta f(X, \theta)].
    \]
\end{lemma}

\begin{lemma}[Jensen's Inequality] \label{lem:jensens}
    Let $X$ be a random variable and $\varphi: \mathbb{R} \to \mathbb{R}$ be a convex function. If $\mathbb{E}[|X|] < \infty$ and $\mathbb{E}[|\varphi(X)|] < \infty$, then the following inequality holds \cite[Theorem 1.6.2]{durrett2019probability}:
    \[
        \varphi(\mathbb{E}[X]) \leq \mathbb{E}[\varphi(X)].
    \]
\end{lemma}

\begin{lemma}[Continuous Mapping Theorem] \label{lem:cmt}
    Let $X_n$ be a sequence of random vectors taking values in $\mathbb{R}^k$, and let $g: \mathbb{R}^k \to \mathbb{R}^m$ be a measurable function. Let $C_g$ be the set of points where $g$ is continuous. If $X_n \xrightarrow{p} X$ (convergence in probability) and $\mathbb{P}(X \in C_g) = 1$, then $g(X_n) \xrightarrow{p} g(X)$. The identical implication holds for convergence in distribution ($\xrightarrow{d}$) and almost sure convergence ($\xrightarrow{a.s.}$) \cite[Theorem 2.3]{vandervaart1998asymptotic}.
\end{lemma}

\begin{lemma}[Portmanteau's Lemma] \label{lem:portmanteau}
    Let $X_n$ and $X$ be random vectors. The sequence $X_n$ converges in distribution to $X$ (denoted $X_n \xrightarrow{d} X$) if and only if for every bounded and continuous function $f$, the following holds \cite[Lemma 2.2]{vandervaart1998asymptotic}:
    \[
        \lim_{n \to \infty} \mathbb{E}[f(X_n)] = \mathbb{E}[f(X)].
    \]
\end{lemma}

\begin{theorem}[Lower bound of MSE for M-estimators]
\label{thm:lowerboundLocalMSE}
Under Assumptions~\ref{ass:1}--\ref{ass:4},
\[
\mathbb{E}[\|\hat{\theta}_n-\theta^*\|_2^2]=\Omega(n^{-1}).
\]
\end{theorem}
\begin{proof}
$X_1, X_2, \ldots, X_n$ are i.i.d. random variables from the data distribution. To prove the statement it is sufficient to lower bound the expected mean squared error $n \mathbb{E}[\|\hat{\theta}_n - \theta^*\|_2^2]$.

From Assumption \ref{ass:1} we have $\theta^* \in \text{int }\Theta$. Since $\theta^*$ minimizes $F_0(\theta) = \mathbb{E}[f(\theta; X)]$ we know $\nabla F_0(\theta^*) = 0$. To apply Lemma~\ref{lem:dct_diff}, define:
\[
g(x):=\|\nabla f(\theta^*;x)\|_2+\rho\bigl(\|\nabla^2 f(\theta^*;x)\|_2+\rho L(x)\bigr).
\]
For $\theta\in U$, Assumption~\ref{ass:3} gives
\[
\begin{aligned}
\|\nabla^2 f(\theta;x)\|_2
&\leq \|\nabla^2 f(\theta^*;x)\|_2+\|\nabla^2 f(\theta;x)-\nabla^2 f(\theta^*;x)\|_2\\
&\leq \|\nabla^2 f(\theta^*;x)\|_2
+L(x)\|\theta-\theta^*\|_2\\
&\leq \|\nabla^2 f(\theta^*;x)\|_2+\rho L(x).
\end{aligned}
\]
Line from $\theta^*$ to $\theta$ is in $U$. Hence:
\[
\nabla f(\theta;x)-\nabla f(\theta^*;x)=\int_0^1\nabla^2 f\bigl(\theta^*+t(\theta-\theta^*);x\bigr)(\theta-\theta^*)\,dt.
\]
Which gives:
\[
\begin{aligned}
\|\nabla f(\theta;x)\|_2
&\leq \|\nabla f(\theta^*;x)\|_2+\|\theta-\theta^*\|_2
\bigl(\|\nabla^2 f(\theta^*;x)\|_2+\rho L(x)\bigr)\\
&\leq \|\nabla f(\theta^*;x)\|_2+\rho\bigl(\|\nabla^2 f(\theta^*;x)\|_2+\rho L(x)\bigr)\\
&=g(x).
\end{aligned}
\]
Moment bounds in Assumption~\ref{ass:3} give:
\[
\mathbb E[g(X)]
\leq G+\rho\bigl(\|\nabla^2F_0(\theta^*)\|_2+H+\rho L\bigr)<\infty.
\]
Hence from Lemma~\ref{lem:dct_diff}, applied on $\operatorname{int}U$, we have:
\begin{equation} \label{eq:pop_grad}
    \boxed{\mathbb{E}[\nabla f(\theta^*; X)] = \nabla F_0(\theta^*) = 0.}
\end{equation}
The displayed Hessian bound is also integrable:
\[
\begin{aligned}
&\mathbb E\bigl[\|\nabla^2 f(\theta^*;X)\|_2+\rho L(X)\bigr]\leq \|\nabla^2F_0(\theta^*)\|_2+H+\rho L<\infty.
\end{aligned}
\]
For $\theta\in\operatorname{int}U$ and $j,k=1,\ldots,d$,
\[
\begin{aligned}
\left| \frac{\partial^2 f(\theta;x)} {\partial\theta_k\,\partial\theta_j} \right|
&\leq \|\nabla^2 f(\theta;x)\|_2\\
&\leq \|\nabla^2 f(\theta^*;x)\|_2+\rho L(x).
\end{aligned}
\]
Hence Lemma~\ref{lem:dct_diff} applied to each component of $\nabla f$ gives:
\[
\begin{aligned}
\frac{\partial^2F_0(\theta)}{\partial\theta_k\,\partial\theta_j}
&=\frac{\partial}{\partial\theta_k}\mathbb E\!\left[\frac{\partial f(\theta;X)}{\partial\theta_j}\right]\\
&=\mathbb E\!\left[\frac{\partial^2 f(\theta;X)}{\partial\theta_k\,\partial\theta_j}\right].
\end{aligned}
\]
Evaluating at $\theta=\theta^*$ gives
\[
\boxed{\mathbb E[\nabla^2 f(\theta^*;X)]=\nabla^2F_0(\theta^*).}
\]

Since $\theta^*$ is strictly in the interior of $\Theta$, the distance to the boundary $\partial \Theta$ is strictly positive. Let $\delta = \min(\rho, \text{dist}(\theta^*, \partial \Theta)) > 0$, where $\rho$ is the radius from Assumption \ref{ass:3}. Let $V = \{ \theta \in \Theta \mid \|\theta - \theta^*\|_2 \leq \delta/2 \}$.
Then, $V \subset \text{int }\Theta$ and $V \subset U$.

We partition the parameter space into two events based on the empirical estimator $\hat{\theta}_n$:
\begin{itemize}
    \item Case A: $\hat{\theta}_n \notin V$. Here:
    \[ \|\hat{\theta}_n - \theta^*\|_2 > \frac{\delta}{2}. \]
    \item Case B: $\hat{\theta}_n \in V$. Because $V \subset \text{int }\Theta$, $\hat{\theta}_n$ is in the interior of the parameter space. Since $F_n$ is differentiable and $\hat{\theta}_n$ minimizes $F_n$, we have $\nabla F_n(\hat{\theta}_n) = 0$.
\end{itemize}
Applying Taylor's Theorem with integral remainder to the vector-valued function $\nabla F_n$, we get:
\[
    0 = \nabla F_n(\hat{\theta}_n) = \nabla F_n(\theta^*) + \left( \int_0^1 \nabla^2 F_n(\theta^* + t(\hat{\theta}_n - \theta^*)) \, dt \right) (\hat{\theta}_n - \theta^*).
\]
Let $H_n = \int_0^1 \nabla^2 F_n(\theta^* + t(\hat{\theta}_n - \theta^*)) \, dt$. Rearranging gives us $\nabla F_n(\theta^*) = - H_n (\hat{\theta}_n - \theta^*)$. Applying the sub-multiplicative property of the spectral norm we get:
\begin{equation} \label{eq:grad_bound}
    \|\nabla F_n(\theta^*)\|_2 \leq \|H_n\|_2 \|\hat{\theta}_n - \theta^*\|_2.
\end{equation}
To upper bound $\|H_n\|_2$, we use the triangle inequality:
\begin{align*}
    \|H_n\|_2 &\leq \|\nabla^2 F_0(\theta^*)\|_2 + \|\nabla^2 F_n(\theta^*) - \nabla^2 F_0(\theta^*)\|_2 \\
    &\quad + \int_0^1 \|\nabla^2 F_n(\theta^* + t(\hat{\theta}_n - \theta^*)) - \nabla^2 F_n(\theta^*)\|_2 \, dt.
\end{align*}
Since $\hat{\theta}_n \in V \subset U$, the line segment connecting $\theta^*$ and $\hat{\theta}_n$ lies entirely in $U$. By the Lipschitz property of the Hessian (Assumption \ref{ass:3}), for any $\theta \in U$, we have $\|\nabla^2 F_n(\theta) - \nabla^2 F_n(\theta^*)\|_2 \leq \frac{1}{n} \sum_{i=1}^n L(X_i) \|\theta - \theta^*\|_2$.
Let $L_n = \frac{1}{n} \sum_{i=1}^n L(X_i)$. The integral term is bounded by $L_n \|\hat{\theta}_n - \theta^*\|_2 \leq L_n \rho$.

Let $M_0 = \|\nabla^2 F_0(\theta^*)\|_2$. By Assumption \ref{ass:2} we know that $F_0$ is twice differentiable, hence $\nabla^2 F_0(\theta^*)$ is a finite $d\times d$ real matrix, hence $M_0 < \infty$. Assumption \ref{ass:2} also gives $M_0\geq \lambda > 0$. Let $D_n = \|\nabla^2 F_n(\theta^*) - \nabla^2 F_0(\theta^*)\|_2$. We define the random sequence:
\[
    Y_n = M_0 + D_n + L_n \rho.
\]
In Case B, $\|H_n\|_2 \leq Y_n$, which by inequality \ref{eq:grad_bound} implies $\|\hat{\theta}_n - \theta^*\|_2 \geq \frac{\|\nabla F_n(\theta^*)\|_2}{Y_n}$.

Combining Cases A and B gives a bound that holds for every $n$:

\begin{align}
    \|\hat{\theta}_n - \theta^*\|_2 &\geq \min\left( \frac{\delta}{2}, \frac{\|\nabla F_n(\theta^*)\|_2}{Y_n} \right) \label{eq:lower_bound_min} \\
    \implies n\|\hat{\theta}_n - \theta^*\|_2^2 &\geq \min\left( n\frac{\delta^2}{4}, \frac{\|\sqrt{n} \nabla F_n(\theta^*)\|_2^2}{Y_n^2} \right). \label{eq:scaled_lower_bound}
\end{align}

Let $W_n = \sqrt{n} \nabla F_n(\theta^*)$. Because $\mathbb{E}[\nabla f(\theta^*; X)] = 0$ and its covariance matrix $K \succ 0$ is finite (Assumption \ref{ass:4}), the multivariate Central Limit Theorem states that:
\[
    W_n \xrightarrow{d} W \sim \mathcal{N}(0, K).
\]
By the Weak Law of Large Numbers, whose required finite first absolute moment condition follows from Assumption~\ref{ass:3} together with Jensen's inequality, since for each relevant quantity \(Y\),
\[\mathbb E|Y|\leq\bigl(\mathbb E|Y|^8\bigr)^{1/8}<\infty\]
whenever Assumption~\ref{ass:3} gives \(\mathbb E|Y|^8<\infty\) we have:
\[D_n \xrightarrow{P} 0\qquad\text{and}\qquad L_n \xrightarrow{P} \mathbb E[L(X)].\] Therefore, by the Continuous mapping theorem:
\[
    Y_n \xrightarrow{P} M_0 + \mathbb{E}[L(X)]\rho \equiv y_0.
\]
Note that $y_0 \geq M_0 \geq \lambda > 0$ by Assumption \ref{ass:2}. Thus, by Slutsky's Theorem:
\[
    Z_n \equiv \frac{\|W_n\|_2^2}{Y_n^2} \xrightarrow{d} \frac{\|W\|_2^2}{y_0^2} \equiv Z.
\]

For any constant $C > 0$, there exists an integer $N$ such that for all $n \geq N$, $n\frac{\delta^2}{4} \geq C$. Therefore, for all $n \geq N$, the inequality \eqref{eq:scaled_lower_bound} implies:
\[
    n\|\hat{\theta}_n - \theta^*\|_2^2 \geq \min(C, Z_n).
\]
Taking expectations on both sides preserves the inequality:
\[
    n \mathbb{E}[\|\hat{\theta}_n - \theta^*\|_2^2] \geq \mathbb{E}[\min(C, Z_n)].
\]
The function $g(x) = \min(C, x)$ is continuous and bounded on $[0,\infty)$ and $Z_n,Z\geq0$. By the Portmanteau's Lemma for weak convergence (since $Z_n \xrightarrow{d} Z$), we have:
\[
    \lim_{n \to \infty} \mathbb{E}[\min(C, Z_n)] = \mathbb{E}[\min(C, Z)].
\]
Since $K \succ 0$, the limiting distribution $W \sim \mathcal{N}(0, K)$ is a non-degenerate multivariate Gaussian, which implies $\mathbb{P}(\|W\|_2 > 0) = 1$. Consequently, $Z > 0$ almost surely, ensuring that:
\[
    c \equiv \mathbb{E}[\min(C, Z)] > 0.
\]
It follows immediately that:
\[
    \liminf_{n \to \infty} n \mathbb{E}[\|\hat{\theta}_n - \theta^*\|_2^2] \geq c > 0.
\]
By definition, this proves that $\mathbb{E}[\|\hat{\theta}_n - \theta^*\|_2^2] = \Omega(1/n)$.
\end{proof}

Combining the bounds of Theorems \ref{thm:boundLocalSquaredBias}, \ref{thm:upperboundLocalMSE}, and \ref{thm:lowerboundLocalMSE} we get the following corollary.

\begin{corollary}
[MSE and Squared Bias Bounds for M-estimators]
\label{cor:M-est}
Under Assumptions~\ref{ass:1}--\ref{ass:5},
\[
\mathbb{E}[\|\hat{\theta}_n-\theta^*\|_2^2]
=\Theta(n^{-1}),
\qquad
\|\mathbb{E}[\hat{\theta}_n-\theta^*]\|_2^2
=\Theta(n^{-2\alpha}).
\]
\end{corollary}

\section{Sample-Size Weighting}
\label{app:weighting}

\begin{theorem}[Optimality of the Server Estimator] \label{thm:optimalnost}
Under Assumptions~\ref{ass:1}--\ref{ass:4}, let $p>0$. Server-only estimation, with $c_0=1$ and $c_1=0$, is order-optimal if and only if $p=\Omega(mn)$.
\end{theorem}

\begin{proof}
\noindent$(\Leftarrow)$ Suppose $p=\Omega(nm)$.
Let $N = nm + p$ be the total amount of data. The parameter $\overline{\theta}$ is considered order-optimal if its MSE is bounded as $\mathbb{E}[\|\overline{\theta}-\theta^*\|_{2}^{2}] = \mathcal{O}(\frac{1}{N})$. We will show this holds for the specified choice of weights.

We choose the weights $c_0=1$ and $c_1=0$. This choice satisfies the condition $mc_1+c_0=1$. With these weights, the weighted average estimator simplifies to:
$$ \overline{\theta} = c_1\sum_{i=1}^{m}\theta_i + c_0\theta_0 = 0\cdot\sum_{i=1}^{m}\theta_i + 1\cdot\theta_0 = \theta_0 $$
From Theorem \ref{thm:upperboundLocalMSE} we have:

$$ \mathbb{E}[\|\overline{\theta}-\theta^*\|_{2}^{2}] =\mathbb{E}[\|\theta_0-\theta^*\|_{2}^{2}] = \mathcal{O}\left(\frac{1}{p}\right). $$

We are given the assumption $p =\Omega(nm)$, therefore the total number of data points $N = nm+p$ is of the same order as $p$:
$$ nm+p =\mathcal{O}(p) \quad \text{and} \quad p =\mathcal{O}(nm+p) \quad \implies \mathcal{O}\left(\frac{1}{p}\right) = \mathcal{O}\left(\frac{1}{nm+p}\right) $$

Hence, the MSE of the estimator $\overline{\theta}$ with $c_0=1$ and $c_1=0$ matches the order-optimal rate:
$$ \mathbb{E}[\|\overline{\theta}-\theta^*\|_{2}^{2}] = \mathcal{O}\left(\frac{1}{N}\right) $$

\medskip
\noindent$(\Rightarrow)$ Suppose that the server-only estimation is order-optimal, i.e.,
\[
\mathbb E[\|\theta_0-\theta^*\|_2^2]
=\mathcal O(N^{-1}).
\]
From the fact that $p$ has a limit we have that either $p\to\infty$ or $p$ is eventually a fixed positive integer.

If $p$ was fixed zero MSE would give us $\theta_0=\theta^*$ a.s. Since $\theta^*\in\operatorname{int}\Theta$, empirical optimality would give $\nabla F_p(\theta^*)=0$ a.s., contradicting Assumption~\ref{ass:4}:
\[
\operatorname{Cov}(\nabla F_p(\theta^*))=\frac{K}{p}\succ0.
\]
This means that for fixed $p$ the server MSE is a positive constant. Since $N=mn+p\to\infty$ we have:
\[
N\mathbb E[\|\theta_0-\theta^*\|_2^2]\longrightarrow\infty,
\]
which is a contradiction with order-optimality assumption. Hence $p\to\infty$.

Theorem~\ref{thm:lowerboundLocalMSE} gives $\mathbb E[\|\theta_0-\theta^*\|_2^2]=\Omega(p^{-1})$. But assumed order-optimality gives $\mathbb E[\|\theta_0-\theta^*\|_2^2]=\mathcal{O}(N^{-1})$. That means:
\[
\frac1p=\mathcal O\!\left(\frac1N\right)
\implies N=\mathcal O(p)
\implies mn=\mathcal O(p)
\implies p=\Omega(mn).
\]
\end{proof}

\begin{proof}[Proof of Theorem~\ref{thm:sample-weighted-strategy}]

\textit{Case 1:} $p\to\infty$.
\begin{align*}
\mathbb{E}[\|\overline{\theta}-\theta^*\|_2^2]&=
\frac{mn^2}{(nm+p)^2}\mathbb{E}[\|\theta_1-\theta^*\|_2^2]+\frac{p^2}{(nm+p)^2}\mathbb{E}[\|\theta_0-\theta^*\|_2^2]\\
&\quad+\frac{m(m-1)n^2}{(nm+p)^2}\|\mathbb{E}[\theta_1-\theta^*]\|_2^2+\frac{2mnp}{(nm+p)^2}\left\langle\mathbb{E}[\theta_1-\theta^*],\mathbb{E}[\theta_0-\theta^*]\right\rangle.
\end{align*}

From Corollary~\ref{cor:M-est} we have:

\begin{align*}
&\frac{mn^2}{(nm+p)^2}\mathbb{E}[\|\theta_1-\theta^*\|_2^2]+\frac{p^2}{(nm+p)^2}\mathbb{E}[\|\theta_0-\theta^*\|_2^2]=\Theta\!\left(\frac{nm+p}{(nm+p)^2}\right)=\Theta\!\left(\frac{1}{nm+p}\right).
\end{align*}

From Theorem~\ref{thm:boundLocalSquaredBias},
\[\frac{m(m-1)n^2}{(nm+p)^2}\|\mathbb{E}[\theta_1-\theta^*]\|_2^2=\Theta\!\left(\frac{m^2}{(nm+p)^2n^{2\alpha-2}}\right).\]

Also,
\begin{align*}
&\left|\frac{2mnp}{(nm+p)^2}\left\langle\mathbb{E}[\theta_1-\theta^*],\mathbb{E}[\theta_0-\theta^*]\right\rangle\right|\overset{\text{C-S}}{\leq}\frac{2mnp}{(nm+p)^2}\|\mathbb{E}[\theta_1-\theta^*]\|_2\|\mathbb{E}[\theta_0-\theta^*]\|_2\\
&\overset{\text{Thm.~\ref{thm:boundLocalSquaredBias}}}{=}\mathcal{O}\!\left(\underbrace{\frac{nm}{nm+p}}_{\leq1}\underbrace{p^{1-\alpha}}_{\leq1}\frac{1}{(nm+p)n^\alpha}\right)\subseteq\mathcal{O}\!\left(\frac{1}{(nm+p)n^\alpha}\right)\subseteq o\!\left(\frac{1}{nm+p}\right).
\end{align*}

This term may be negative, but its absolute value is $o((nm+p)^{-1})$ so it cannot cancel the positive lower bound from the first two terms. Combining these results we get:
\[
\mathbb{E}[\|\overline{\theta}-\theta^*\|_2^2]=\Theta\!\left(\frac{1}{nm+p}+\frac{m^2}{(nm+p)^2n^{2\alpha-2}}\right).
\]

\textit{Case 2:} $p\geq1$ is fixed. By Assumption~\ref{ass:1},
\[
\mathbb E[\|\theta_0-\theta^*\|_2^2]\leq R^2,\qquad\|\mathbb E[\theta_0-\theta^*]\|_2\leq R.
\]
Since $N=mn+p=\Theta(mn)$, the same expansion gives:
\begin{align*}
\frac{mn^2}{N^2}\mathbb E[\|\theta_1-\theta^*\|_2^2]&=\Theta\!\left(\frac{mn}{N^2}\right)=\Theta(N^{-1}),\\
\frac{p^2}{N^2}\mathbb E[\|\theta_0-\theta^*\|_2^2]&\leq\frac{p^2R^2}{N^2}=\mathcal O(N^{-2}),\\
\frac{m(m-1)n^2}{N^2}\|\mathbb E[\theta_1-\theta^*]\|_2^2&=\Theta\!\left(\frac{m^2}{N^2n^{2\alpha-2}}\right),\\
\left|\frac{2mnp}{N^2}\left\langle\mathbb E[\theta_1-\theta^*],\mathbb E[\theta_0-\theta^*]\right\rangle\right|&\leq\frac{2mnpR}{N^2}\|\mathbb E[\theta_1-\theta^*]\|_2\\
&=\mathcal O\!\left(\frac{1}{Nn^\alpha}\right)=o(N^{-1}).
\end{align*}
Hence,
\[
\mathbb E[\|\overline\theta-\theta^*\|_2^2]
=\Theta\!\left(\frac1N+\frac{m^2}{N^2n^{2\alpha-2}}\right).
\]

\textit{Case 3:} $p=0$. Then $c_0=0$, $c_1=1/m$, and $N=mn$. Independence gives
\begin{align*}
\mathbb E[\|\overline\theta-\theta^*\|_2^2]&=\frac1m\mathbb E[\|\theta_1-\theta^*\|_2^2]+\frac{m-1}{m}\|\mathbb E[\theta_1-\theta^*]\|_2^2\\
&=\Theta\!\left(\frac1{mn}+\frac1{n^{2\alpha}}\right)=\Theta\!\left(\frac1N+\frac{m^2}{N^2n^{2\alpha-2}}\right).
\end{align*}

In all three cases, the MSE is $\mathcal{O}((nm+p)^{-1})$ if and only if
\[
\frac{m^2}{(nm+p)^2n^{2\alpha-2}}=\mathcal{O}\!\left(\frac{1}{nm+p}\right) \iff \frac{m^2}{n^{2\alpha-2}}=\mathcal{O}(nm+p).
\]
\end{proof}

\begin{proof}[Proof of Corollary~\ref{cor:sample-weighted-regimes}]
If $m=\mathcal{O}(n^{2\alpha-1})$, then
\[
\frac{m^2}{n^{2\alpha-2}}=\mathcal O(nm)
\implies
\frac{m^2}{n^{2\alpha-2}}=\mathcal O(nm+p).
\]
Hence by Theorem~\ref{thm:sample-weighted-strategy} the first statement is true.

If $m=\omega(n^{2\alpha-1})$, then
\[
\frac{nm}{m^2/n^{2\alpha-2}}=\frac{n^{2\alpha-1}}{m}\longrightarrow0.
\]
Hence,
\[
\frac{m^2}{n^{2\alpha-2}}=\mathcal{O}(nm+p)\iff\frac{m^2}{n^{2\alpha-2}}=\mathcal{O}(p)\iff p=\Omega\!\left(\frac{m^2}{n^{2\alpha-2}}\right).
\]

Hence by Theorem~\ref{thm:sample-weighted-strategy} the second statement is true.

\end{proof}

\begin{proof}[Proof of Corollary~\ref{cor:gap}]
\[
p=\Omega(nm)\overset{\text{Thm.~\ref{thm:optimalnost}}}{\implies}\text{server-only weighting achieves order-optimal MSE}.
\]
From $p=o(nm\cdot\frac{m}{n^{2\alpha-1}})$ we have $p\neq\Omega\!\left(nm\cdot\frac{m}{n^{2\alpha-1}}\right)$. Hence:
\[
m=\omega(n^{2\alpha-1}) \land p\neq\Omega\!\left(nm\cdot\frac{m}{n^{2\alpha-1}}\right)\overset{\text{Cor.~\ref{cor:sample-weighted-regimes}}}{\implies}\hspace{2mm}
\begin{gathered}
\text{sample-size weighting does not}\\
\text{achieve order-optimal MSE}.
\end{gathered}
\]
\end{proof}

\section{Analytical Example Calculations}
\label{app:examples}

\subsection{An unbiased sample mean}

For the model in Section~\ref{sec:unbiased_mean},
\[
F_0(\theta)=\tfrac12\bigl((\theta-q)^2+q(1-q)\bigr),
\qquad
F_n'(\theta)=\theta-\bar X_n.
\]
Thus $\theta^*=q$ and $\hat\theta_n=\bar X_n\in[0,1]\subset\Theta=[-1,2]$. Since $\mathbb E[\bar X_n]=q$ and $\operatorname{Var}(\bar X_n)=q(1-q)/n$, the estimator is unbiased. Sample-size weighting gives
\[
\overline\theta
=\frac1N\left(\sum_{i=1}^m\sum_{j=1}^n X_{ij}
+\sum_{j=1}^p X_{0j}\right)
=\hat\theta_{\mathrm{pool}},
\qquad
\mathbb E[(\overline\theta-\theta^*)^2]=\frac{q(1-q)}{N}.
\]
The server sum is omitted when $p=0$.

Assumption~\ref{ass:1} holds because $\Theta$ is compact and convex, with $q$ in its interior. The sample Hessian and the population Hessian both equal $1$. The gradient $\theta-X$ is bounded on $\Theta$, and the Hessian is constant. This verifies Assumptions~\ref{ass:2} and~\ref{ass:3}. At $\theta^*=q$,
\[
K=\mathbb{E}[(q-X)^2]
=(1-q)q^2+q(q-1)^2
=q(1-q)>0,
\]
so Assumption~\ref{ass:4} also holds. Assumption~\ref{ass:5} fails because the bias is zero.

\subsection{Misspecified least squares}

For the model in Section~\ref{sec:misspecified_least_squares},
\[
F_0'(\theta)=\mathbb E[Z^2]\theta-\mathbb E[Z^4]
=\frac52\theta-\frac{17}{2},
\qquad
F_n'(\theta)=\frac1n\sum_{j=1}^n(\theta Z_j^2-Z_j^4).
\]
Consequently,
\[
\theta^*=\frac{17}{5},
\qquad
\hat\theta_n=\frac{\sum_{j=1}^n Z_j^4}{\sum_{j=1}^n Z_j^2}.
\]
The empirical minimizer is a weighted average of the values $Z_j^2$, with positive weights proportional to $Z_j^2$. It therefore lies in $[1,4]\subset\Theta=[0,5]$, so the constraint is inactive.

The gradient is $\theta Z^2-ZY$, and the Hessian is $Z^2$. The population Hessian equals $5/2>0$. The data are bounded, and the sample Hessian does not depend on $\theta$. At $\theta^*=17/5$, the gradient takes the values $12/5$ and $-12/5$ with equal probability. Its mean is zero, so
\[
K=\frac12\left(\frac{12}{5}\right)^2
+\frac12\left(-\frac{12}{5}\right)^2
=\frac{144}{25}>0.
\]
These facts verify Assumptions~\ref{ass:1}--\ref{ass:4}.

We calculate the bias by a Taylor expansion. This is the usual method for approximating moments of a smooth function of a sample mean; see Oehlert~\cite{oehlert1992}. Write
\[
\overline{Z^2}_n=\frac1n\sum_{i=1}^n Z_i^2.
\]
Since $Z_i^4=5Z_i^2-4$, the estimator becomes
\[
\hat{\theta}_n=5-\frac{4}{\overline{Z^2}_n}.
\]
The variables $Z_i^2-5/2$ are independent and take the values $-3/2$ and $3/2$ with equal probability. Thus
\[
\mathbb{E}\!\left[\overline{Z^2}_n\right]=\frac52,
\qquad
\mathbb{E}\!\left[
\left(\overline{Z^2}_n-\frac52\right)^2
\right]
=\frac{9}{4n}.
\]
In the fourth moment, only products with all four indices equal or with two pairs of equal indices have nonzero expectation. Any other product contains an index that appears only once. Independence then gives a factor with expectation zero.

There are $n$ terms with all four indices equal. Each has expectation $(3/2)^4=81/16$. For two pairs of equal indices, there are $\binom{n}{2}$ choices of distinct indices. Each pair can be arranged in the four positions in $\binom{4}{2}=6$ ways. For $i\neq j$, independence gives
\[
\mathbb{E}\!\left[
\left(Z_i^2-\frac52\right)^2
\left(Z_j^2-\frac52\right)^2
\right]
=
\mathbb{E}\!\left[\left(Z_i^2-\frac52\right)^2\right]
\mathbb{E}\!\left[\left(Z_j^2-\frac52\right)^2\right]
=\left(\frac94\right)^2.
\]
This gives
\begin{align*}
\mathbb{E}\!\left[
\left(\overline{Z^2}_n-\frac52\right)^4
\right]
&=\frac1{n^4}
\mathbb{E}\!\left[
\left(\sum_{i=1}^n\left(Z_i^2-\frac52\right)\right)^4
\right]\\
&=\frac1{n^4}\Biggl[
\sum_{i=1}^n
\mathbb{E}\!\left[\left(Z_i^2-\frac52\right)^4\right]\\
&\qquad+
6\sum_{1\leq i<j\leq n}
\mathbb{E}\!\left[
\left(Z_i^2-\frac52\right)^2
\left(Z_j^2-\frac52\right)^2
\right]
\Biggr]\\
&=\frac1{n^4}
\left[
n\frac{81}{16}
+6\binom{n}{2}\frac{81}{16}
\right]\\
&=\frac{81}{16n^4}\bigl[n+3n(n-1)\bigr]\\
&=\frac{81(3n-2)}{16n^3}.
\end{align*}

Expanding $5-4/\overline{Z^2}_n$ around $\overline{Z^2}_n=5/2$ gives
\[
\hat{\theta}_n
=\frac{17}{5}
+\frac{16}{25}\left(\overline{Z^2}_n-\frac52\right)
-\frac{32}{125}\left(\overline{Z^2}_n-\frac52\right)^2
+R_n.
\]
The absolute value of the third derivative of this expression with respect to its argument is at most $24$ on $[1,4]$. Since $\overline{Z^2}_n\in[1,4]$, Taylor's remainder satisfies
\[
|R_n|
\leq4\left|\overline{Z^2}_n-\frac52\right|^3.
\]
By Cauchy--Schwarz,
\begin{align*}
\mathbb{E}\!\left[
\left|\overline{Z^2}_n-\frac52\right|^3
\right]
&\leq
\sqrt{
\mathbb{E}\!\left[
\left(\overline{Z^2}_n-\frac52\right)^2
\right]
\mathbb{E}\!\left[
\left(\overline{Z^2}_n-\frac52\right)^4
\right]
}\\
&=O(n^{-3/2}).
\end{align*}
Hence $\mathbb{E}[|R_n|]=O(n^{-3/2})$. Taking expectations in the expansion gives
\[
\mathbb{E}[\hat{\theta}_n-\theta^*]
=-\frac{32}{125}\frac{9}{4n}+O(n^{-3/2})
=-\frac{72}{125n}+O(n^{-3/2}).
\]
All five assumptions hold, with $\alpha=1$ and $b=-72/125$. This is the bias relative to the best linear predictor, $\theta^*=17/5$. Squared loss can therefore give either an unbiased estimator or a biased one.

\subsection{Squared loss with a quartic penalty}

The sample gradient and Hessian are
\[
\frac{\partial f}{\partial\theta}(\theta;X)
=\theta+\theta^3-X,
\qquad
\frac{\partial^2 f}{\partial\theta^2}(\theta;X)
=1+3\theta^2.
\]
Since $\mathbb{E}[X]=0$, the population derivative is $F_0'(\theta)=\theta+\theta^3$. Its unique zero is $\theta^*=0$, which lies in the interior of $\Theta$. The Hessian is at least $1$. This verifies Assumptions~\ref{ass:1} and~\ref{ass:2}.

For the gradient, $|\theta|\leq1$ and $|X|\leq1$ give
\[
\left|\frac{\partial f}{\partial\theta}(\theta;X)\right|
=|\theta+\theta^3-X|
\leq|\theta|+|\theta|^3+|X|
\leq3.
\]
Its eighth absolute moment is therefore at most $3^8$.

For any $\theta,\theta'\in[-1,1]$, the Hessian satisfies
\begin{align*}
\left|
\frac{\partial^2 f}{\partial\theta^2}(\theta';X)
-\frac{\partial^2 f}{\partial\theta^2}(\theta;X)
\right|
&=3|(\theta')^2-\theta^2|\\
&=3|\theta'-\theta|\,|\theta'+\theta|\\
&\leq6|\theta'-\theta|.
\end{align*}
It is therefore $6$-Lipschitz on $\Theta$. The constant $6$ also satisfies the moment conditions on the Lipschitz constant in Assumption~\ref{ass:3}.

The Hessian does not depend on $X$, so the population Hessian is also $F_0''(\theta)=1+3\theta^2$. Thus the centered Hessian is
\[
\frac{\partial^2 f}{\partial\theta^2}(\theta;X)-F_0''(\theta)
=(1+3\theta^2)-(1+3\theta^2)=0.
\]
Its eighth absolute moment is zero. This verifies Assumption~\ref{ass:3}.

At $\theta^*=0$, the gradient equals $-X$. Since $\mathbb{E}[X]=0$,
\[
K=\operatorname{Var}(-X)
=\operatorname{Var}(X)
=\mathbb{E}[X^2]-(\mathbb{E}[X])^2
=\mathbb{E}[X^2].
\]
If $\mathbb{E}[X^2]=0$, then $X=0$ almost surely. This would imply $\mathbb{E}[X^3]=0$, contrary to our choice of distribution. Hence $K>0$, which verifies Assumption~\ref{ass:4}.

Write $\overline{X}_n=n^{-1}\sum_{i=1}^n X_i$. The empirical derivative is strictly increasing. At the endpoints it satisfies $F_n'(-1)=-2-\overline{X}_n<0$ and $F_n'(1)=2-\overline{X}_n>0$. The empirical minimizer therefore lies in $(-1,1)$ for every sample and satisfies
\[
\hat{\theta}_n+\hat{\theta}_n^3=\overline{X}_n.
\]
This gives $|\hat{\theta}_n|\leq|\overline{X}_n|$ and $\overline{X}_n-\hat{\theta}_n=\hat{\theta}_n^3$. Factoring the difference of cubes yields
\begin{align*}
\left|\hat{\theta}_n-\overline{X}_n+\overline{X}_n^3\right|
&=\left|\overline{X}_n^3-\hat{\theta}_n^3\right|\\
&=|\overline{X}_n-\hat{\theta}_n|\,
\left|
\overline{X}_n^2+\overline{X}_n\hat{\theta}_n+\hat{\theta}_n^2
\right|\\
&\leq3|\overline{X}_n|^5.
\end{align*}

Independence and $\mathbb{E}[X]=0$ imply
\[
\mathbb{E}[\overline{X}_n]=0,
\qquad
\mathbb{E}[\overline{X}_n^3]
=\frac{n\mathbb{E}[X^3]}{n^3}
=\frac{\mathbb{E}[X^3]}{n^2}.
\]
In the cubic expansion, only terms with all three indices equal have nonzero expectation.

To bound the fifth absolute moment, expand the sixth power of $\sum_{i=1}^n X_i$. A product has expectation zero if an index appears only once. The remaining products contain at most three distinct indices. There are $O(n^3)$ such products, and each has absolute value at most one. Dividing by $n^6$ gives $\mathbb{E}[\overline{X}_n^6]=O(n^{-3})$. Since the function $t\mapsto t^{5/6}$ is concave on $[0,\infty)$, Jensen's inequality gives
\begin{align*}
\mathbb{E}[|\overline{X}_n|^5]
&=\mathbb{E}\!\left[(\overline{X}_n^6)^{5/6}\right]\\
&\leq
\bigl(\mathbb{E}[\overline{X}_n^6]\bigr)^{5/6}.
\end{align*}
The sixth-moment bound means that $\mathbb{E}[\overline{X}_n^6]\leq Cn^{-3}$ for some constant $C>0$ and all sufficiently large $n$. Hence
\[
\mathbb{E}[|\overline{X}_n|^5]
\leq(Cn^{-3})^{5/6}
=C^{5/6}n^{-5/2}
=O(n^{-5/2}).
\]

The absolute value of an expectation is at most the expectation of the absolute value. The difference-of-cubes bound therefore gives
\begin{align*}
\left|
\mathbb{E}[\hat{\theta}_n]
-\mathbb{E}[\overline{X}_n]
+\mathbb{E}[\overline{X}_n^3]
\right|
&\leq
\mathbb{E}\!\left[
\left|\hat{\theta}_n-\overline{X}_n+\overline{X}_n^3\right|
\right]\\
&\leq3\mathbb{E}[|\overline{X}_n|^5]\\
&=O(n^{-5/2}).
\end{align*}
Substituting $\mathbb{E}[\overline{X}_n]=0$ and $\mathbb{E}[\overline{X}_n^3]=\mathbb{E}[X^3]/n^2$ gives
\[
\left|
\mathbb{E}[\hat{\theta}_n]
+\frac{\mathbb{E}[X^3]}{n^2}
\right|
=O(n^{-5/2}).
\]
Since $\theta^*=0$, this is equivalent to
\[
\mathbb{E}[\hat{\theta}_n-\theta^*]
=-\frac{\mathbb{E}[X^3]}{n^2}+O(n^{-5/2}).
\]
All five assumptions hold, with
\[
\alpha=2,
\qquad b=-\mathbb{E}[X^3]\neq0.
\]
For the concrete distribution in Section~\ref{sec:quartic_example},
\[
\mathbb E[X^3]
=\frac23\left(-\frac13\right)^3
+\frac13\left(\frac23\right)^3
=\frac{2}{27},
\qquad b=-\frac{2}{27}.
\]

\section{Numerical Experiment Details}
\label{app:numerics}

\subsection{Synthetic calculations}
The synthetic experiments in Section~\ref{sec:numerics} use the three models of Section~\ref{sec:concrete_examples}, with $q=1/3$ for the Bernoulli mean and $\mathbb P(X=-1/3)=2/3$, $\mathbb P(X=2/3)=1/3$ for the quartic example. Each local estimator is determined by a binomial count: the number of ones for the sample mean, the number of observations with $Z=2$ for least squares, and the number with $X=2/3$ for the quartic model. The corresponding success probabilities are $1/3$, $1/2$, and $1/3$.

We sum over these counts to compute the finite-sample moments of the local estimators. Binomial probabilities are evaluated using probability ratios around the mode. A Hoeffding bound selects a retained range with total omitted probability at most $10^{-18}$; retained probabilities are renormalized. For the quartic model, twelve Newton iterations solve $\hat\theta_n+\hat\theta_n^3=\bar X_n$. The identity $\mathbb E[\hat\theta_n]=-\mathbb E[\hat\theta_n^3]$ reduces cancellation in its small bias. The least-squares bias calculation likewise uses the symmetry of its binomial distribution. The unbiased mean uses its exact bias and variance.

Independence and the weights $c_0=p/N$, $c_1=n/N$ then determine the aggregate MSE, including the worker--server bias cross term. Server-only and pooled MSE are obtained from the same local-moment calculations at sample sizes $p$ and $N$. Thus the curves use numerical finite-sample expectations, rather than Monte Carlo averages or substitution into the asymptotic rate formula.

The grid is $n\in\{8,16,32,64,128,256\}$. The three models first use $m=n^2$, $p=n^3$. The remaining least-squares regimes are $(m,p)=(n,n)$, $(n^2,\lceil n^{7/2}\rceil)$, and $(n^2,n^4)$. Separate Monte Carlo checks use 20,000 repetitions at $n=8,16$ in all six regimes; they do not supply the plotted curves. The plotted quantity is $N\operatorname{MSE}$, and the reference slopes are the theoretical predictions.

\subsection{CIFAR-10 data and representation}
The experiments condition on the fixed pool of 50,000 CIFAR-10 training images, the frozen feature extractor, and the 10,000-image test set. Worker and server observations are drawn independently with replacement from the training pool. The same draws are used across aggregation strategies and classifier heads within a repetition. As $m$ changes, the same ordered worker sample is repartitioned into equal-sized local datasets. As $p$ changes, server datasets are nested prefixes of a separately sampled stream. Repetitions use independent training draws. The test set is used only for evaluation.

The extractor is Torchvision's ResNet-18 with \texttt{IMAGENET1K\_V1} weights and the final classification layer replaced by the identity. Its evaluation transforms resize images to 256 pixels on the shorter side, apply a central $224\times224$ crop, and use ImageNet channel normalization. Each resulting 512-dimensional feature vector is normalized to unit Euclidean norm. The extractor remains in evaluation mode and is not trained on CIFAR-10. No data augmentation is used.

\subsection{Local fitting and aggregation}
Ridge classifiers minimize the mean half squared error against one-hot labels plus $0.01/2$ times the squared Frobenius norm of the coefficient matrix. They have no intercept. The penalty coefficient is fixed in the average-loss objective at every sample size. Fits use a direct linear-system solve in double precision, using the primal or dual system according to its dimension. Ridge parameter MSE is the mean squared Frobenius distance to the ridge minimizer over the complete training pool, without division by parameter dimension. This reference is distinct from the pooled estimator fitted to the sampled $N$ observations in each repetition.

The linear cross-entropy head is affine. The nonlinear head has a $512\to128\to10$ architecture with a ReLU hidden layer and biases in both layers. Both use mean cross-entropy plus $10^{-4}/2$ times the squared Euclidean norm of all parameters, including biases, as the training objective. Each fit uses 400 full-batch Adam updates with learning rate $0.01$. When observations are processed in chunks, gradients are accumulated over the complete local dataset before an update. Pooled fits use the same optimization budget.

Within each head and repetition, workers, server, and pooled fits start from identical parameters. The linear head starts at zero. Nonlinear weights are drawn from zero-mean Gaussian distributions with variances $2/512$ and $1/128$ in the first and second layers, respectively; biases start at zero. Nonlinear initializations vary across repetitions. Sample-size weighting averages corresponding fitted parameters, including both nonlinear layers and their biases. Common initialization does not guarantee that learned hidden units remain aligned. There is no further optimization after aggregation.

\subsection{Experimental grids and uncertainty}
The fixed-budget study uses $mn=32768$ and $p=8192$. Ridge uses $m\in\{2,4,8,16,32,64,128,256,512\}$; the cross-entropy heads use $m\in\{2,8,32,128,512\}$. The server-size study uses $mn=8192$, $m\in\{8,128,512\}$, and $p\in\{128,512,2048,8192,32768,131072\}$. The main-text server-size panels report $m=128$. There are 20 repetitions for ridge and 10 for each cross-entropy head. Sampling and initialization use base seed 20261022 with separate random streams.

Reported confidence intervals are approximate pointwise 95\% Student-$t$ intervals for mean metrics over repetitions. Their half-width is the corresponding Student-$t$ quantile times the sample standard deviation divided by the square root of the repetition count. Comparisons between methods use differences within the same repetition. These intervals describe variation conditional on the fixed training pool, feature extractor, and test set.

Accuracy is the fraction of correctly classified test images. Only ridge is assigned parameter MSE. The saved prediction-loss metric is one-half the mean sum of squared output errors against one-hot labels: it uses raw ridge scores and softmax probabilities for the cross-entropy heads. It is distinct from parameter MSE and, for the cross-entropy heads, from the training loss.

The longer-training check uses the first three repetitions of the fixed-budget study at $m=8,512$, comparing 400 and 800 updates for both cross-entropy heads on the same data and initialization. This diagnostic does not certify convergence or global optimality. No bias exponent or verification of Assumption~\ref{ass:5} is claimed for CIFAR-10; the nonlinear comparison lies outside the convex theory.

\fi

\end{document}